\documentclass[letterpaper, 10 pt, conference]{ieeeconf}  

\IEEEoverridecommandlockouts                              

\usepackage{graphics}  
\usepackage{epsfig}    
\usepackage{amsmath}   
\usepackage{bm}
\usepackage{mathtools}
\usepackage{commath}
\usepackage{color}
\usepackage{soul}
\usepackage{amssymb}
\usepackage{comment}
\usepackage{xfrac}
\usepackage{nicefrac}
\usepackage[ruled,vlined]{algorithm2e}
\usepackage{verbatim}
\usepackage[font=small]{caption}
\usepackage[font=small]{subcaption}

\usepackage[percent, abs]{overpic}
\usepackage{hyperref}

\usepackage{xcolor,varwidth}
\usepackage{graphicx}
\usepackage[export]{adjustbox}

\usepackage{tikz}

\usepackage[noadjust]{cite}
\usepackage{filecontents}

\usepackage{placeins}

\usepackage{float}
\usepackage{gensymb}
\usepackage{multirow}

\usepackage{algorithmicx}
\usepackage{algcompatible}

\usepackage{xcolor,colortbl}
\usepackage{dblfloatfix}

\renewcommand{\baselinestretch}{0.97}

\title{\LARGE \bf
Minimal Footprint Grasping Inspired by Ants
}

\author{Mohamed Sorour and Barbara Webb
\thanks{Institute for Perception, Action and Behaviour,
School of Informatics, University of Edinburgh. Informatics Forum, 10 Crichton St, EH8 9AB Edinburgh, United Kingdom.
{\tt\small msorour@ed.ac.uk}}%
}

\begin{document}

\maketitle
\thispagestyle{empty}
\pagestyle{empty}

\begin{abstract}
Ants are highly capable of grasping objects in clutter, and we have recently observed that this involves substantial use of their forelegs. The forelegs, more
specifically the tarsi, have high friction microstructures (setal pads), are covered in hairs, and have a flexible under-actuated tip. Here we abstract these features to test their functional advantages for a novel low-cost gripper design, suitable for bin-picking applications. In our implementation, the gripper legs are long and slim, with high friction gripping pads, low friction hairs and single-segment tarsus-like structure to mimic the insect's setal pads, hairs, and the tarsi's interactive compliance. Experimental evaluation shows this design is highly robust for grasping a wide variety of individual consumer objects, with all grasp attempts successful.  In addition, we demonstrate this design is effective for picking single objects from dense clutter, a task at which ants also show high competence. The work advances grasping technology and shed new light on the mechanical importance of hairy structures and tarsal flexibility in insects.

\end{abstract}

\section{Introduction}

The need to grasp in tight spaces and cluttered environments arises frequently in robot grasping applications. Examples include bin/shelf picking \cite{Correll2016,Fujita2020,Yu2016,Gustavo2020,Hernandez2017}, conveyor belt segregation \cite{Nguyen_2024,Han_2020}, laboratory automation \cite{Harazono_2024}, agri-robotics \cite{Eduardo_2023}, and kitting \cite{Drigalski_2020} to name only a few. These applications dictate a gripper design with minimal interacting element foot-print. Consequently, most of the current gripper designs used in practical applications and grasping challenges feature suction pads and/or two fingered grippers \cite{Chen2015,Morrison2017,Fujita2019}. 
Suction is most effective for light objects with flat surfaces but needs to be augmented with other technologies in hybrid designs to deal with a wider range of items, as well as requiring an external, typically bulky, system (e.g. a compressor) to create the suction. Two fingered grippers often require specialised adaptations to fit specific industrial needs \cite{Wang2021,Kobayashi2019,Guo2017}. The capabilities of insects such as ants to handle a wide variety of objects in a large range of scenarios suggest that a more general solution should be possible.

\begin{figure}[t!]
\centering
\includegraphics[width=0.8\linewidth]{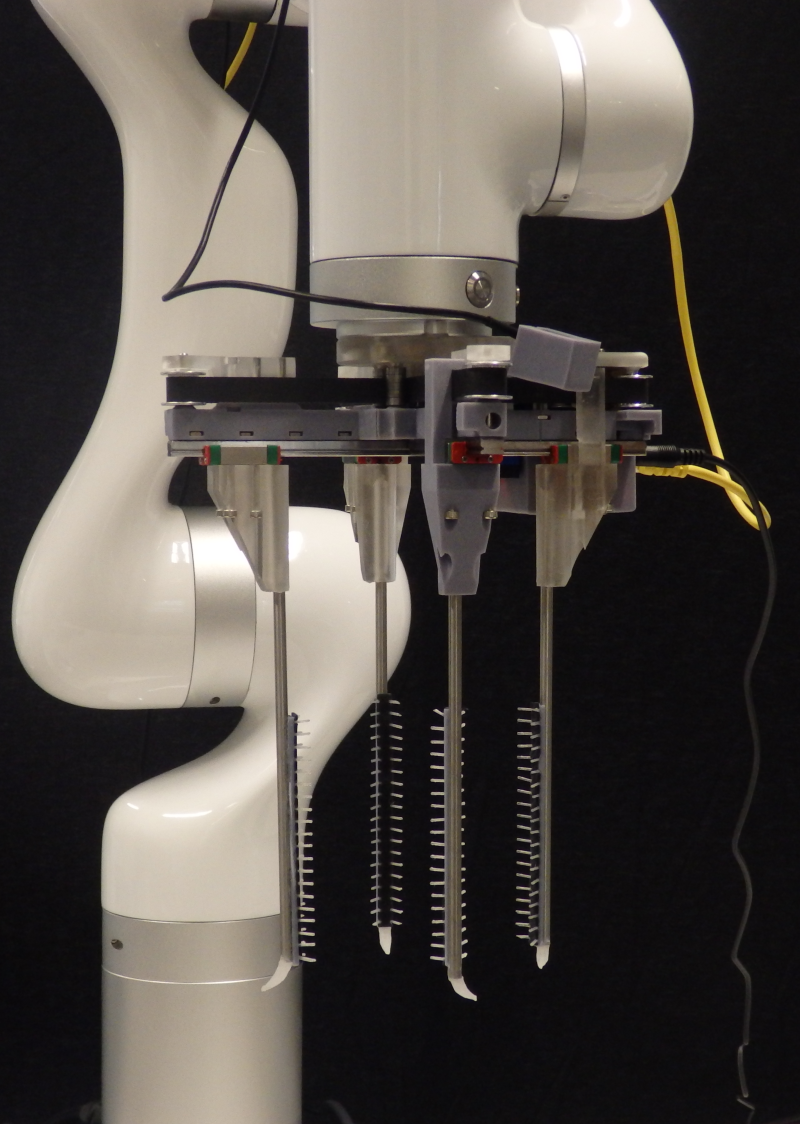}
\captionsetup{aboveskip=10pt}
\captionsetup{belowskip=-15pt}
\caption{A novel gripper design mimicking the functional capabilities of the ant tarsus}
\label{figure_gripper_setup}
\end{figure}

\begin{figure}[t!]
\centering
\includegraphics[width=1.\linewidth]{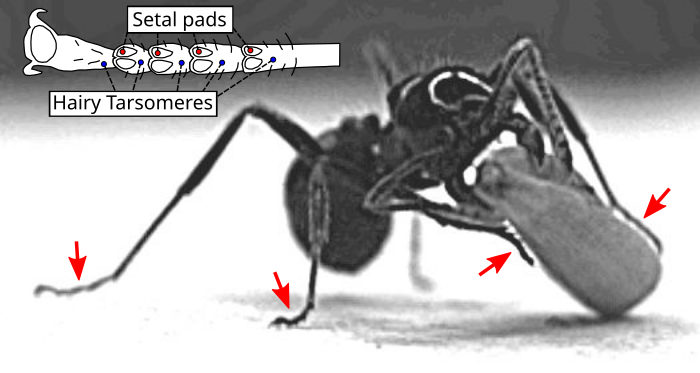}
\captionsetup{aboveskip=10pt}
\captionsetup{belowskip=-15pt}
\caption{Ants use foreleg's tarsus to grasp and manipulate objects.}
\label{figure_insect_leg_anatomy}
\end{figure}

Insect-inspired robotic designs for grasping are recently gaining interest. Some are inspired by the ant's mandible biting mechanism \cite{Zhang2020,Wipfler2024}, including the mechanical advantage of the hairs that cover the inner surfaces of the mandibles \cite{Sorour2025_AntGrip}. Others have focused on the leg tarsi, shown in the schematic in Fig. \ref{figure_insect_leg_anatomy}. The tarsus is primarily a passive element, conforming and adhering via cuticular elasticity and microstructures (pulvilli, tenent hairs) \cite{Chapman_2012}, while active control is restricted to tendon-driven claw movement. Insect tarsi have inspired research in grasping, locomotion and climbing robots. In \cite{Julian_2024}, a robotic gripper is presented, consisting of multiple under-actuated tendon driven digits resembling the tarsal chain. In \cite{ZHAO_2023}, researchers developed flexible biomimetic friction pads mimicking the structure of insect euplantulae (located on the ventral side of an insect's tarsus) found in grasshoppers, locusts, and cockroaches. In \cite{Phodapol_2023}, inspired by the tarsus segments of hornets, a compliant, underactuated robotic gripper was developed that mimics the insect’s tendon-driven mechanism to wrap around and grasp objects. A hybrid attachment system combining claws and adhesive pads was introduced in \cite{Song_2016}, revealing that their synergistic interaction enhances adhesion on rough surfaces more effectively than either component alone, offering insights for climbing robot design. Additional studies have explored the beetle’s exceptional grip on cylindrical surfaces \cite{Dagmar_2017}, the critical role of dual pre-tarsus claws in robust attachment \cite{Gorb_2025}, and the evolutionary adaptation of euplantulae’s adhesive microstructures \cite{Buscher_2018}. The spring-like elasticity observed in crane fly and mosquito joints has informed innovations in flying robot landing systems \cite{Hyun_2025}, robotic leg design \cite{Tran_Ngoc_2022}, and miniature tree-climbing robots \cite{Ishibashi_2022}. Notably, \cite{Judith_2022} emphasized that both euplantula microstructures (for smooth surfaces) and claws (for mechanical interlocking on uneven terrain) are indispensable for insects to achieve stable locomotion and attachment across diverse environments.

The insect leg plays an important role in grasping as well as locomotion, and the tarsus section is pivotal. In most insects, this is subdivided into two to five tarsomeres. These are differentiated from leg segments by the absence of muscles \cite{Chapman_2012}. The adhesive organs are diverse, including pulvilli, and tarsal setal pads, which are composed of thousands of tiny, flexible hair-like structures. The flexibility and high number of these setae allow them to conform to irregular surfaces, greatly enhancing adhesion --- usually studied in the context of climbing but equally important for grasping \cite{Chapman_2012}. Additional flexibility of the tarsi structure stems from being a multi-joint, tendon-driven chain that remains passive pre-grasping (with inherent joint spring-elasticity in some insects \cite{Frazier_1999}), enabling it to explore and adapt to an object's contours, a critical feature given the insect's poor close-range eyesight. In insects that perform frequent grasp activities, common morphological adaptations include one- or two-segmented tarsi and a single claw \cite{Chapman_2012}. To date, inspiration from insect tarsi for robotics, such as the combination of adhesive pads and claws, has been explored largely in the domain of irregular objects with rough surfaces naturally present in the insect's habitat. The potential application to the man-made smooth, regular industrial packaging and designs has not been explored.

In this study, rather than closely copying the structural details of insect legs, we focused on replicating some of their key functional principles.
Specifically, we replicated three major elements: first, mimicking the adhesive microstructures of euplantulae using a thin, high-friction, elastic material; second, replicating the tarsus's pre-grasp flexibility through a flexible, single-segment interactive component that facilitates gripper-object interactions, termed the tarsus-like structure in the following; and third, integrating low-friction, hair-like structures to separate the high-friction pads from the object during the pre-grasp phase. Once the gripper is positioned, these structures flexibly deform under grasping force, enabling the friction material to make contact with the target surface. In combination, these elements allow the legs themselves to be long and slim without compromising grasp success and stability, with a minimal footprint that facilitates grasping of objects in clutter.

The paper is organized as follows: section II presents a novel, four-leg gripper design and provides a stress analysis to demonstrate its feasibility. Section III describes successful experiments for grasping a wide variety of items and grasping in clutter. Conclusions are finally given in section IV.

\section{Insect Leg Gripper}

\begin{figure}[t!]
\centering
\includegraphics[width=1.0\linewidth]{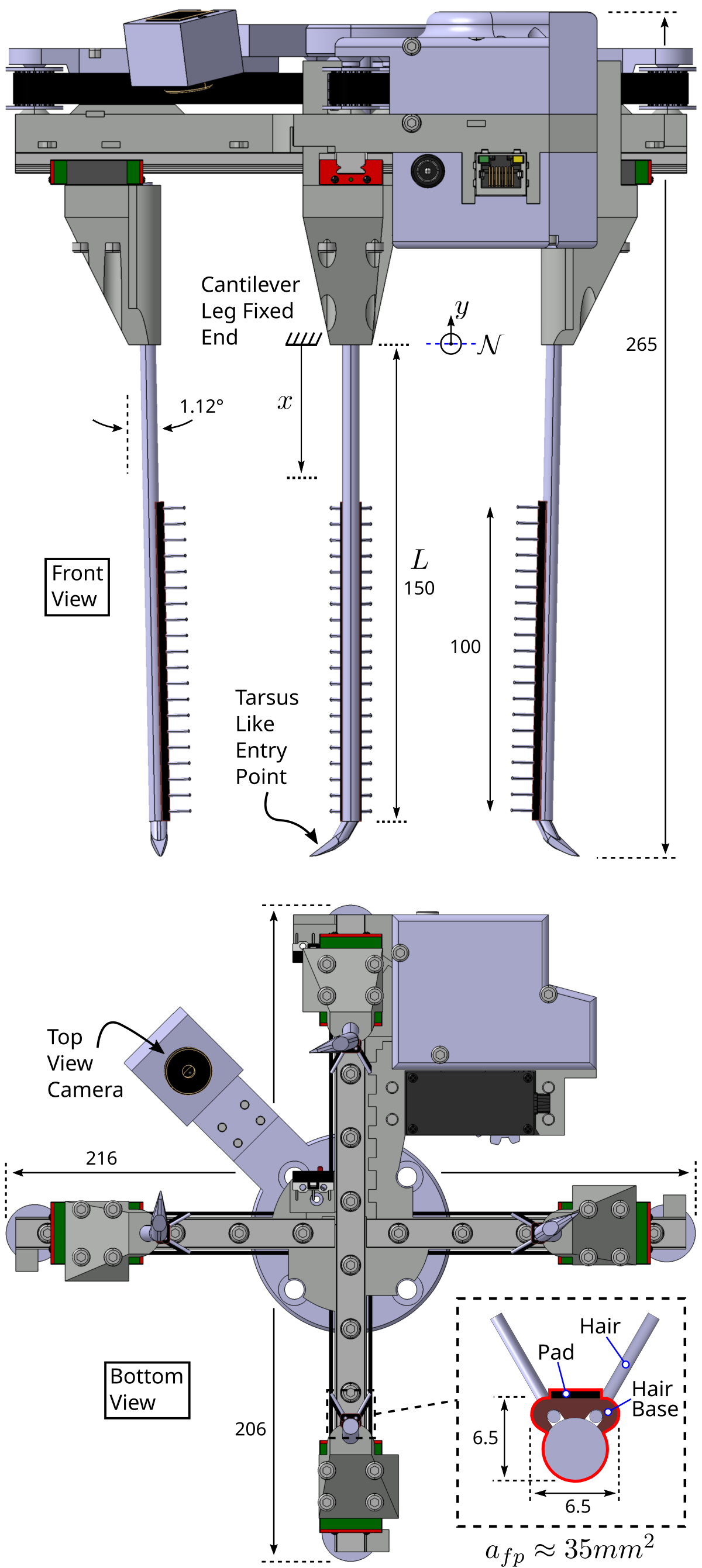}
\captionsetup{belowskip=-10pt}
\captionsetup{aboveskip=10pt}
\caption{Insect leg gripper CAD model in front and bottom views, key dimensions shown in millimeters.}
\label{figure_gripper_CAD_model}
\end{figure}

The CAD model of the designed gripper is illustrated in Fig. \ref{figure_gripper_CAD_model}. The gripper features four insect-inspired legs equipped with inward-facing hairs and terminated by a single-segment tarsus-like structure. Each leg incorporates a high-friction pad 0.6 mm thick with protruding hairs angled at 60°, identified in prior research \cite{Sorour2025_AntGrip} to optimize support for grasped objects. In this design, the hairs serve an additional purpose: they keep the friction pads isolated from the object during pre-grasp positioning in dense clutter. The friction pads, highlighted in black within the dotted box of Fig. \ref{figure_gripper_CAD_model}(bottom view), are constructed from F80 elastic resin \cite{F80_resin} with a shore hardness of 50-60A, while the hairs are made of low-friction thermoplastic polyurethane (TPU). The hair base is fabricated from hard resin instead of the elastic material to prevent tearing at the hair attachment points under repeated grasping forces. 

The legs themselves are made from polished stainless steel, ensuring a low-friction outer surface that minimizes disturbance when navigating through cluttered environments. Each leg measures 150 mm in length, with the friction pad and hairs covering only the lower 100 mm. At the end of each leg, a tarsus-like structure protrudes at an arbitrary angle relative to the grasping surface. Its irregular shape, featuring a decreasing cross-section and curved design akin to insect tarsi, prevents pressure buildup upon impact with objects or supporting surfaces. These are also made from TPU material, which provides a balance between  rigidity for pushing between objects and flexibility for safe interaction with surfaces while retaining its form post-deformation. 

Actuation is obtained through timing belt transmission, with each leg secured to one side of the same belt as shown in Fig. \ref{figure_gripper_CAD_model}(bottom view), introducing limited passive compliance and ensuring simplicity due to a single degree of freedom (DoF) to actuate all legs. Linear bearings/rails constrain each leg’s motion. The gripper’s opening ranges from 5 mm (minimum) to 120 mm (maximum), with a positioning accuracy of ±1 mm, suitable for the intended application. 

A camera mounted on the gripper provides a top view for recording and analyzing interactions with objects. The gripper has overall dimensions of 206×216×265 mm and weighs 760 grams. It operates in either position or force control modes: position control regulates the opening gap, while force control closes the fingers until a specified force is achieved, measured via calibrated actuator current. We introduce two metrics, maximum $a_{fp}$ and total $A_{fp}$ gripper footprint areas, to assess a gripper's ability to penetrate and grasp objects in densely cluttered environments. The former, represented by the red-shaded area within the dashed box in Fig. \ref{figure_gripper_CAD_model} (bottom), refers to the maximum cross-sectional area of the gripper's leg/finger end-tip that interacts with and penetrates an object clutter. Lower values of $a_{fp}$ are preferable for effective performance in such scenarios. In our design, all legs have a uniform $a_{fp}$ of 35 mm², calculated conservatively by excluding the tarsus-like structure. The total footprint area $A_{fp}$ sums to 140 mm² and represents the total in-between object-clutter gap area necessary for the gripper to overcome in the pre-grasp phase. A lower value is again better, and as such a lower number of gripper legs/fingers.

\subsection{Stress Analysis}

This gripper design is aimed at grasping general household objects. Through prior experience, the force required to grasp these can be up to $50N$. For that we have identified the gripper legs as a critical element of the design for which we perform stress analysis to formulate the leg's deflection under load and make sure it is within the elastic limit of the material used. The analysis is also essential in the design cycle to obtain a minimal leg diameter and in turn gripper penetration footprint. The gripper leg is modeled as a cantilever rod of length $L$, the fixed end of which is shown in Fig. \ref{figure_gripper_CAD_model} (top), subjected to a point load $F$ at its free end. The cross sectional area at any point $x$ along the cantilever length is circular in shape with diameter $D$. The bending stress at any cross section along the beam is given by \cite{budynas2015shigleys}:
\begin{equation}
    \sigma(x,y)= \frac{M(x).y}{I(x)},
\end{equation}
where $M(x)=F.(L-x)$ is the bending moment computed as the applied force multiplied by the arm length to the cross section, $y$ is the distance from the neutral axis $\mathcal{N}$ shown in Fig. \ref{figure_gripper_CAD_model}(top) in blue, and $I=\pi.D^4/64$ is the area moment of inertia of the circular section. Since we are interested in the maximum stress value, occurring at the outer fiber of the cross section, we can substitute with $y=D/2$. The maximum value of stress at a given cross section is as such formulated by:
\begin{equation}
    \sigma(x)= \frac{32F.(L-x)}{\pi.D^3}.
    \label{eq_bending_stress}
\end{equation}
For a uniform cross section cantilever, the maximum stress is always at the fixed end (at $x=0$), therefore the maximum stress value $\sigma_{max}$:
\begin{equation}
    \sigma_{max}= \frac{32F.L}{\pi.D^3}.
    \label{eq_bending_stress_max}
\end{equation}
On the other hand, the beam deflection $\delta_x$ at any point $x$ from the fixed end is given by \cite{budynas2015shigleys}:
\begin{equation}
    \delta_x= \frac{F.x^2}{6E.I}(3L-x),
    \label{eq_deflection}
\end{equation}
with $E$ being the Young's modulus of the finger material. For the respective cantilever beam the maximum deflection $\delta_{x_{max}}$ value is always at the free end i.e $x=L$:
\begin{equation}
    \delta_{x_{max}}= \frac{64F.L^3}{3\pi.E.D^{4}}.
    \label{eq_deflection_max}
\end{equation}
As a nominal operation example for the gripper iterative design, we assume a grasping force of 50N applied through 3 gripper legs. For that the point load $F$ for each leg is computed as 16.67N, we also assume an average object height of 40mm, leading to a bending moment arm length $L$ of 120mm (discussed further below). For a stainless steel leg material with $E=200$ GPa and yield strength of $200$ MPa, the smallest (commercially available) leg diameter $D$ evaluates to 5mm. Based on these design parameters, the maximum bending stress in Eq.(\ref{eq_bending_stress_max}) evaluates to $163$ MPa well below the material's yield strength of $200$ MPa at which structural deflections become plastic (permanent deformations) by $18.5\%$ margin. The leg deflection at such grasping force evaluates as $\delta_{x_{max}}=1.56mm$ from Eq.(\ref{eq_deflection_max}).
To get the slope of the deflection at the free end (slope at maximum deflection) we compute the derivative of (\ref{eq_deflection}) with respect to $x$ and then substitute with $x=L$ to get:
\begin{equation}
    \theta_{max}= \frac{32F.L^2}{\pi.E.D^{4}}.
    \label{eq_slope_of_deflection_max}
\end{equation}
Evaluating Eq.(\ref{eq_slope_of_deflection_max}) with the parameters above, we obtain a maximum leg bending slope of $1.12\degree$ which is used for an inward leg inclination as depicted in Fig. \ref{figure_gripper_CAD_model} to maximize leg-object grasping contact and to ensure the object is pushed inward rather than out of the gripper.

\begin{figure}[t!]
\centering
\includegraphics[width=1.0\linewidth]{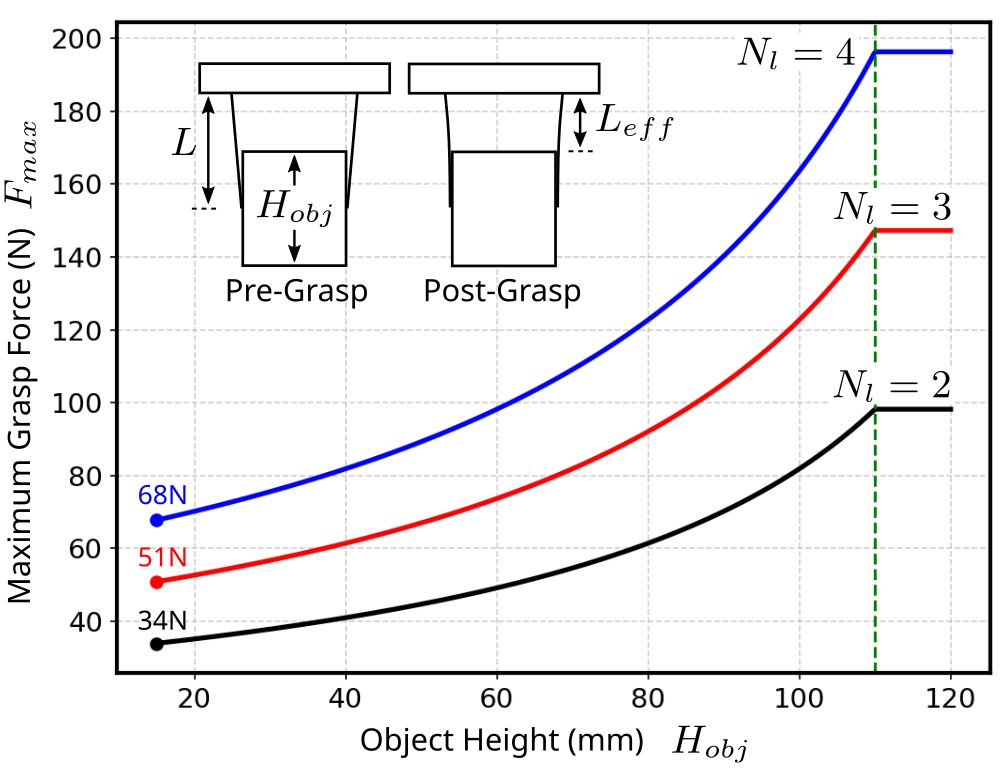}
\captionsetup{aboveskip=10pt}
\captionsetup{belowskip=-15pt}
\caption{Maximum applicable grasping force versus object height.}
\label{figure_max_grasp_force}
\end{figure}

\begin{figure*}[t!]
\centering
\includegraphics[width=1.0\textwidth]{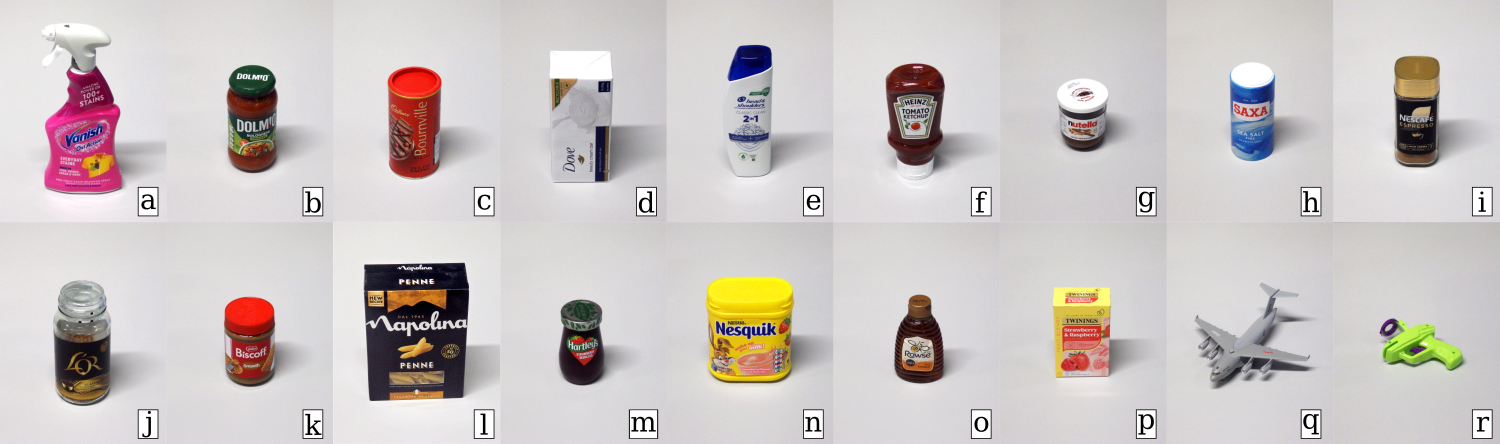}
\captionsetup{belowskip=-10pt}
\captionsetup{aboveskip=5pt}
\caption{Set of objects used to evaluate the proposed gripper. (a) Sprayer packaging (490gm), (b) sauce glass bottle (690gm), (c) chocolate drink cylinder (295gm), (d) soap bar pack (600gm), (e) shampoo (375gm), (f) ketchup bottle (500gm), (g) nutella (375gm), (h) saxa salt package (384gm), (i) nescafe coffee jar (323gm), (j) lor coffee jar (442gm), (k) biscoff spread (610gm), (l) pasta box (533gm), (m) jam jar (500gm), (n) nesquick drink (452gm), (o) honey bottle (365gm), (p) drink teabags (60gm), (q) toy1 (150gm), and (r) toy2 (45gm).}
\label{figure_set_of_objects}
\end{figure*}

\subsection{Maximum grasping force}

The maximum applicable grasping force by the gripper without excessive leg bending depends firstly on the shape of the object, where regular shapes as cylinders, squares, and rectangles can have four point contacts with equally distributed leg loads. Non regular objects on the other hand will have at least two uniformly distributed leg loads (in worst case scenario at equilibrium state) and up to four non-uniformly distributed leg loads. The maximum force depends secondly on the height of the object and in turn on the bending moment arm length $L$ in Eq.(\ref{eq_bending_stress_max}) and Eq.(\ref{eq_deflection_max}), represented by the variable $x$ in Fig. \ref{figure_gripper_CAD_model}. The leg length is 150 mm, the load bearing high friction part of which is 100 mm long, and assuming a 10 mm distance between the rigid free end of the leg to the object-resting surface (to accommodate for the tarsus-like flexible structure), the minimum and maximum arm length $L$ is 50 mm and 145 mm respectively. The latter value assumes a minimum object height $H_{obj}$ of 15 mm. Due to the initial inward leg bending, during grasp execution, what starts as a point leg-object contact, eventually turns into a non-uniformly distributed line contact. How long is the line contact depends on the amount of exerted force, however, here we are interested in the maximum applicable force, so we will assume a line contact covering the whole object height. This means that the arm length is no longer a fixed value $L$, but dependent on the object height, we designate such varying value as the effective arm length $L_{eff}$ evaluated as follows:
\begin{equation}
    L_{eff} = \max \left( 0.05,\; 0.16 - H_{obj} \right)
    \label{eq_effective_arm_length}
\end{equation}
Solving for the force $F$ in Eq.(\ref{eq_bending_stress_max}), substituting the stress value $\sigma_{max}$ with the material’s yield strength $\sigma_{yield}$ (the absolute maximum stress the material can handle before plastic deformation occurs), and the arm length $L$ with the effective value $L_{eff}$, we obtain the maximum applicable grasping force $F_{max}$:
\begin{equation}
    F_{max}=\frac{\pi.N_{l}.D^3.\sigma_{yield}}{32\max(0.05,0.16-H_{obj})},
    \label{eq_max_gripper_force}
\end{equation}
where $N_{l}$ is the number of engaged gripper legs in grasping. The maximum grasping force evaluated using Eq.(\ref{eq_max_gripper_force}) is shown in Fig. \ref{figure_max_grasp_force} for object heights varying from 15mm to 120mm, and for the possible number of gripper legs engaged. It is worth noting that Eq. (\ref{eq_max_gripper_force}) represented in this figure applies to both regular and irregular shaped objects despite there being no way to predict where the contact will happen with the latter without grasp planning. In Fig. \ref{figure_max_grasp_force} the lowest value of each curve is indicated, for example, the maximum force value in case of two leg grasping (black curve) of an object 15mm high is 34N. This demonstrates that for objects with regular geometries, much higher grasping forces can be achieved. This indicates the potential for employing slimmer gripper legs (4mm diameter), thereby reducing the overall footprint in grasping applications involving exclusively such shapes.


\section{Experiments}
Two sets of experiments have been conducted. The first aims at grasping individual objects and manipulating these to test the grasp robustness. The second set aims at picking an object from a cluttered bin to show the penetration advantage of the minimal footprint gripper legs.

\begin{figure*}[b!]
\centering
\includegraphics[width=1.0\textwidth]{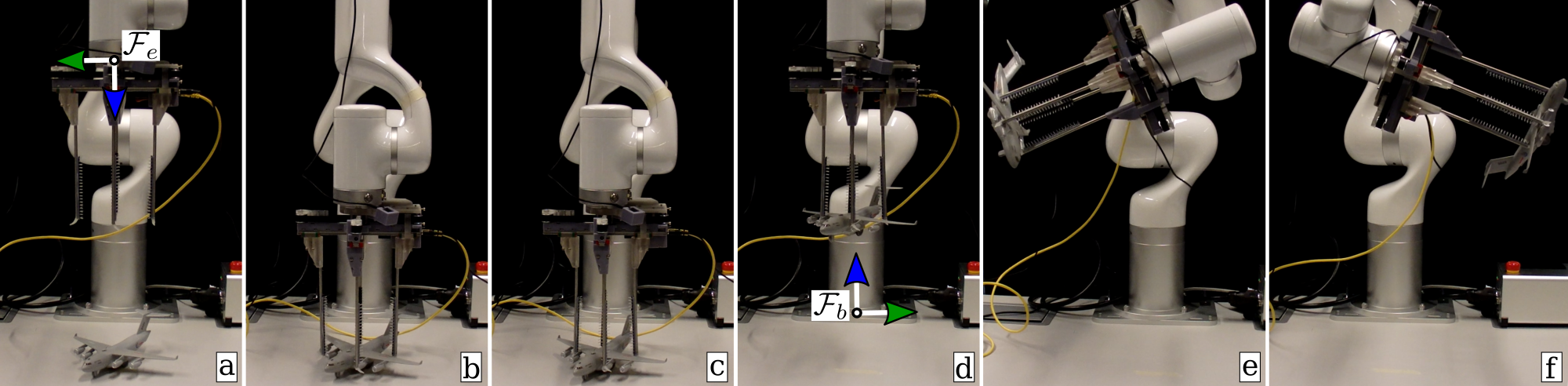}
\captionsetup{belowskip=-16pt}
\captionsetup{aboveskip=5pt}
\caption{Snapshots of the steps in a first experiment set, starting with home pose (a), then descending (b) grasping (c) and lifting (d), manipulating to test the grasp stability (e)-(f).}
\label{figure_experiment1_snapshots}
\end{figure*}

\subsection{Setup and Methodology}
The developed gripper is fitted to the collaborative robot uFactory xArm 850, as shown in Fig. \ref{figure_gripper_setup}, capable of manipulating a payload of 5 Kg with 850 mm reach. Both the robot arm and the gripper are controlled in \texttt{Python} by sending 6D pose, and position/force commands respectively. In a first set of experiments, the 18 objects depicted in Fig. \ref{figure_set_of_objects} are grasped individually, each is placed on a fixed mark on the table, corresponding to the gripper centerline at the arm home pose, shown in Fig. \ref{figure_experiment1_snapshots}(a). 
The set of objects tested include cylindrical/circular shapes with varying radii, which are best suited for the 4-leg gripper structure (and typically challenging for 2-finger or suction grippers), as well as objects with flat, symmetrical sides, and some irregularly shaped objects. The objects were otherwise selected to provide variety in texture and deformability. These weigh from 45 to 690 grams, with 399 grams on average, high enough to excite the object dynamics during the manipulation phase. For each object, 1 to 3 grasp attempts were executed depending on how challenging the object geometry was (manually evaluated). Grasping is followed by an abrupt manipulation phase to test grasp robustness.

\begin{figure*}[t!]
\centering
\includegraphics[width=1.0\textwidth]{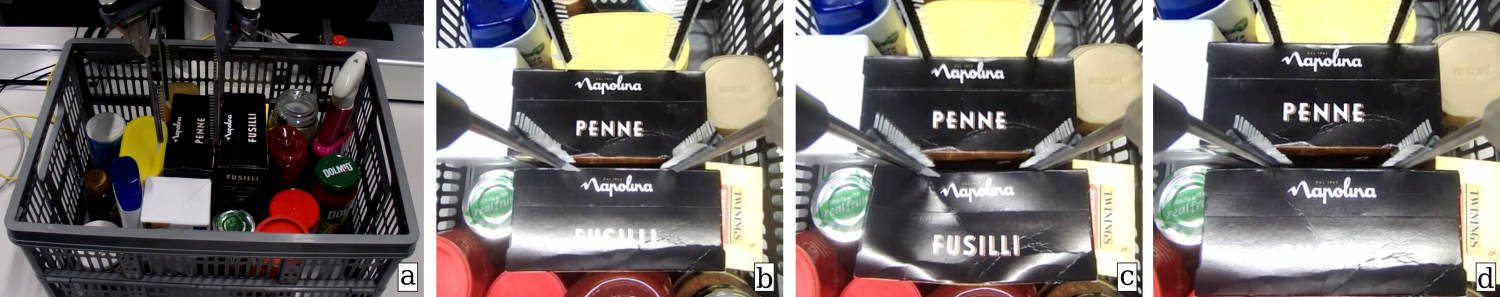}
\captionsetup{belowskip=-16pt}
\captionsetup{aboveskip=5pt}
\caption{Snapshots of grasping an object amoung clutter in a second experiment, starting with home pose (a), then adjusting gripper opening and moving to pre-grasp pose (b), descending with tarsus-like structure interacting with object (c), followed by grasping (d) and lifting.}
\label{figure_experiment2_snapshots}
\end{figure*}

Snapshots of the robot movement sequence in this experiment are presented in Fig. \ref{figure_experiment1_snapshots}. A very similar sequence was used in evaluating an earlier ant-mandible inspired gripper \cite{Sorour2025_AntGrip}.  Each grasp trial consists of 9 consecutive steps, made up of 7 arm movements and 2 gripper actions, starting with the arm in the home pose $\textbf{\texttt{P}}_\texttt{{HOME}}$, Fig. \ref{figure_experiment1_snapshots}(a), and the gripper in a predefined opening gap depending on the width of the object to be grasped. The first three steps attempt to grasp and lift the object:
\begin{enumerate}
    \item The arm descends to the pose $\textbf{\texttt{P}}_\texttt{{GRASP}}$, Fig. \ref{figure_experiment1_snapshots}(b).
    \item The gripper closes in force control mode until the $15N$ command force is achieved, Fig. \ref{figure_experiment1_snapshots}(c).
    \item The arm then lifts back to the $\textbf{\texttt{P}}_\texttt{{HOME}}$, Fig. \ref{figure_experiment1_snapshots}(d).
\end{enumerate}
To test the grasp stability/robustness, the arm then manipulates the grasped object into 2 successive poses, with relatively high acceleration to excite external forces/torques on the object simulating a worst case manipulation scenario:
 \begin{enumerate}
  \setcounter{enumi}{3}
    \item A combination of translatory and rotary motion ending in pose $\textbf{\texttt{P}}_\texttt{{ROT1}}$, Fig. \ref{figure_experiment1_snapshots}(e).
    \item The opposite combination of translatory and rotary motion ending in pose $\textbf{\texttt{P}}_\texttt{{ROT2}}$, Fig. \ref{figure_experiment1_snapshots}(f).
\end{enumerate}
The final steps deposit and release the object (omitted from Fig. \ref{figure_experiment1_snapshots} for redundancy): 
\begin{enumerate}
 \setcounter{enumi}{5}
 \item The arm moves back to the $\textbf{\texttt{P}}_\texttt{{HOME}}$,
 \item The arm descends to the $\textbf{\texttt{P}}_\texttt{{RELEASE}}$ (here identical to the $\textbf{\texttt{P}}_\texttt{{GRASP}}$),
 \item The gripper opens back to the initial gap distance.
 \item Finally the arm terminates at the $\textbf{\texttt{P}}_\texttt{{HOME}}$.
\end{enumerate}
The arm's built-in motion profile parameters are set to a maximum velocity, and acceleration of 400 $mm/s$, and 700 $mm/s^2$ respectively. Let the pose vector $\textbf{\texttt{P}} = \begin{bmatrix*} ^{b}\textbf{r}_{e}^{\top} & \!\!\! ^{b}\bm{\theta}_{e}^{\top} \end{bmatrix*}^{\top} \!\!\! \in \mathbb{R}^{6}$ of the manipulator's end effector frame $\bm{\mathcal{F}}_{e}$ expressed in the base frame $\bm{\mathcal{F}}_b$, shown in Fig. \ref{figure_experiment1_snapshots}(a) and Fig. \ref{figure_experiment1_snapshots}(d) respectively, define the task space coordinates. With $^{b}\textbf{r}_{e} = \begin{bmatrix*} x & y & z \end{bmatrix*}^{\top}$ being the position vector, and $^{b}\bm{\theta}_{e} = \begin{bmatrix*} \alpha & \beta & \gamma \end{bmatrix*}^{\top}$ denoting a minimal representation of orientation (\textit{roll, pitch, yaw} RPY variation of Euler angles), the aforementioned poses are given by:
\begin{flalign*}
\notag &\hspace{25pt} \textbf{\texttt{P}}_\texttt{{HOME}} = \begin{bmatrix*} 0.37 & 0 & 0.5 & 180\degree & 0 & 0 \end{bmatrix*}^{\top},\\ 
\notag &\hspace{25pt} \textbf{\texttt{P}}_\texttt{{GRASP}} = \begin{bmatrix*} 0.37 & 0 & 0.27 \sim 0.35 & 180\degree & 0 & 0 \end{bmatrix*}^{\top},\\
\notag &\hspace{25pt} \textbf{\texttt{P}}_\texttt{{ROT1}} = \begin{bmatrix*} 0.37 & 0 & 0.5 & 180\degree & 70\degree & 90\degree \end{bmatrix*}^{\top},\\ 
\notag &\hspace{25pt} \textbf{\texttt{P}}_\texttt{{ROT2}} = \begin{bmatrix*} 0.37 & 0 & 0.5 & 180\degree & -70\degree & 90\degree \end{bmatrix*}^{\top},
\end{flalign*}
where the position, and orientation values presented in meters, and radians respectively. Note the $z$ coordinate value of $\textbf{\texttt{P}}_\texttt{{GRASP}}$ ranges from 0.27 m to 0.35 m depending on the height of the object to be grasped.

In a second set of experiments, the gripper is tasked with retrieving two objects from the cluttered bin depicted in Fig. \ref{figure_experiment2_snapshots}(a). The first object, a large box in close contact with a neighboring item of similar size, is selected to evaluate the tarsus-like leg-tip’s ability to penetrate clutter, snapshots of which are depicted in Fig. \ref{figure_experiment2_snapshots}. The second object, a relatively short item, is chosen to highlight the benefits of slender, long legs for grasping a short object relative to its neighboring items in cluttered environments. The manipulator begins from the home position, $\textbf{\texttt{P}}_\texttt{{HOME}}$, after which the gripper orientation is aligned with the target object. The arm then moves to the grasp location, and the gripper opening is adjusted to match the object’s size. Subsequently, the arm descends, the gripper closes under force control, and the object is lifted. All poses and gripper openings in this experiment are pre-programmed to illustrate the gripper’s capability for cluttered-scene grasping, rather than to demonstrate perception-based grasp planning, which lies beyond the scope of this work.

\subsection{Results and Discussion}

\begin{figure}[b!]
\centering
\includegraphics[width=1.0\linewidth]{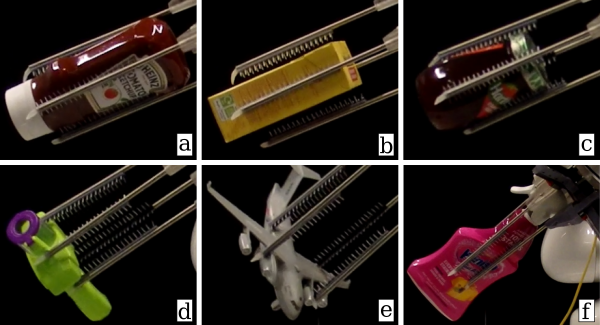}
\captionsetup{belowskip=-16pt}
\captionsetup{aboveskip=5pt}
\caption{Selection of challenging objects.}
\label{figure_particular_objects}
\end{figure}
All the objects depicited in Fig. \ref{figure_set_of_objects} were grasped and manipulated successfully. We have not included a grasp success rate table as there were no failures; the grasping trials for this experiment are fully video documented and available online\footnote{\url{https://drive.google.com/drive/folders/1bYZQ9P\_Hx83o1WUL9y2z65I2GKmraaJa?usp=sharing}.}.


A subset of particularly challenging objects to is shown in Fig. \ref{figure_particular_objects}. The ketchup bottle in Fig. \ref{figure_particular_objects}(a) was not reliably grasped in previous reported work \cite{Sorour2025_AntGrip} due to an inverted taper shape. Here, thanks to the inward inclined legs, and a complete caging due to four point contact, it was always grasped firmly in every grasp iteration. The rectangular-shaped object in Fig. \ref{figure_particular_objects}(b), shows only two gripper legs engaged in grasping, illustrating the versatility of the gripper. Although this gripper is perfectly suited for circular objects, the jam jar in Fig. \ref{figure_particular_objects}(c) is another example of the advantageous inward tilted legs, since it will always have a point rather than line-contact due to its unique geometry. The irregular shaped toys in Fig. \ref{figure_particular_objects}(d), (e) provide another example of versatility of the proposed gripper, where some of the legs deform more than others to maintain multi-point contact to a certain limit. The sprayer in Fig. \ref{figure_particular_objects}(f) is particularly difficult item to grasp, due to a tall low friction surface and a $3D$ complex geometry that always led to unstable point contact with a parallel plate gripper \cite{Sorour2025_AntGrip}. However here, employing the proposed gripper achieved $100\%$ grasp success rate thanks to a forced high friction four point contact.

Larger/heavier rectangular objects, as in Fig. \ref{figure_set_of_objects}(d)(l) as well as irregular objects as in Fig. \ref{figure_set_of_objects}(a) are grasped in between the gripper legs at $45\degree$. An example of this is depicted in Fig. \ref{figure_experiment2_snapshots}. Although the applied grasp force is halved in these grasps (due to applied force at 45 degrees to the surface tangent), the grasps reported are quite robust and successful. Such a grasping configuration, however, allows for grasping large objects as long as one dimension fits into the gripper opening.

To grasp from a clutter, the gripper footprint should be minimal, and this is one of the key design considerations of the gripper. In addition, the tarsus-resembling structure aims to lower the initial contact footprint further and guide the object into the gripper legs upon interaction. This is demonstrated in the second set of experiments and depicted in Fig. \ref{figure_experiment2_snapshots}(c). The TPU tarsus-inspired-tip deforms and allows for a smaller entry point in-between cluttered objects to guide the gripper legs into grasp position. The hairs deform sufficiently to allow the legs to be inserted between closely positioned objects while minimizing any contact of the high friction pads; the latter only come into effect when the gripper closes on the object. Videos of picking both objects are available online \footnote{\url{https://drive.google.com/drive/folders/1AEx7MLAe3aQJCiF5yw1a\_wm50VEz9sxJ?usp=sharing}.}

\section{Conclusion}
In this work, a novel, insect-leg inspired gripper is presented. The developed gripper features four thin legs with slim, high friction pads, separated from objects by low friction hairs in pre-grasp phase suitable for space-tight applications. In addition a low friction single-segment tarsus-like structure helps the gripper interact with and guide objects in clutter penetration. Experimental evaluation on a pool of objects varying in shape shows that the deflection tolerant legs are very useful in adapting to irregular shapes. The results show that insect tarsi can inspire successful grasping devices and that its hairs can also function as a separating medium between the high-friction setal pads and the objects of interest to grasp.

\section*{ACKNOWLEDGMENTS}
This work is funded by EPSRC under grant agreement EP/V008102/1, An insect-inspired approach to robotic grasping.


\bibliographystyle{IEEEtran}  
\bibliography{Bib}

\end{document}